\documentclass[conference]{IEEEtran}
\IEEEoverridecommandlockouts
\usepackage{indentfirst}
\usepackage{booktabs}
\usepackage{epsfig}
\usepackage{cite}
\usepackage{amsmath,amssymb,amsfonts}
\usepackage{algorithmic}
\usepackage{algorithm}
\usepackage{graphicx}
\usepackage{textcomp}
\usepackage{xcolor}
\usepackage{comment}
\usepackage{tabularx}

\def\BibTeX{{\rm B\kern-.05em{\sc i\kern-.025em b}\kern-.08em
    T\kern-.1667em\lower.7ex\hbox{E}\kern-.125emX}}
\begin{document}

\title{Learning to Fine-tune Foundation Models under Resource Limitations}
\author{Thomas Tsouparopoulos and Iordanis Koutsopoulos
\\ Department of Informatics \\
Athens University of Economics and Business\\Athens, Greece}

\maketitle

\begin{abstract}

We study the problem of optimal continual fine-tuning for a pre-trained Foundation Model deployed at a resource-limited device. At each time slot, a new batch of training data arrives, and the controller is faced with two options: either use the data to fine-tune the model and incur a compute cost, or do not fine-tune the model and discard the data. After the decision, the performance of the current model is measured in terms of an application-specific performance metric such as classification accuracy. Our objective is to learn an optimal policy that determines \emph{when to fine-tune the model} on a single task (e.g., sentiment analysis), under a finite compute budget. We formulate this online decision-making problem as a constrained Markov Decision Process, where the system state captures three essential aspects: (\textit{i}) model's performance, (\textit{ii}) computational budget, and (\textit{iii}) data distribution relevance to historic data encountered up to that point. The transition to the next state is stochastic and therefore, we propose a reinforcement learning-based method to solve this problem, namely the \emph{actor-critic} algorithm. We also consider the special case where the performance of fine-tuning for a given model can be predicted or estimated prior to decision; in this case the problem becomes a Dynamic Programming one. Experiments with a large pre-trained model on a widely-used text classification dataset demonstrate that our method consistently outperforms fine-tuning approaches with the same compute budget by more than $4\%$ in terms of accuracy and achieves $97\%$ of full-parameter fine-tuning accuracy while requiring only $25\%$ of the fine-tuning steps.
\end{abstract}


\section{Introduction}

Foundation Models (FMs) are large models pre-trained on massive, general-purpose datasets through self-supervised learning, so that they learn general patterns, logical structures, and relationships between concepts. This process creates a flexible general-purpose \emph{base model} that can then be fine-tuned with smaller datasets for various downstream tasks such as text summarization and code generation, without the need to be retrained from scratch each time. Fine-tuning (FT) involves updating some or all of the model parameters by performing some training iterations with the new, small dataset.

A representative example is a traffic-analysis FM at a 6G base station, initially trained on generic network traces and periodically adapted using locally observed traffic patterns such as new application behaviors or emerging protocol variants. The proposed RL controller learns when to trigger updates based on expected network-level benefit, enabling resource-aware model adaptation that improves traffic classification accuracy while respecting the operational constraints of wireless infrastructure.

Parameter-efficient fine-tuning (PEFT) methods, e.g., Low Rank Adaptation ($\mathrm{LoRA}$)~\cite{hu2022lora} and adapters are the predominant class of approaches for continually updating large FMs by significantly reducing  the number of trainable parameters, allowing for lightweight model updates, while maintaining near full fine-tuning performance at a fraction of the cost. However, the benefit of PEFT methods for a given model changes over time with data distribution shift, and it depends on the history of FT steps applied to the model's parameters~\cite{Aggarwal24}.

When a FM resides on a resource-limited device, the problem of FT the \emph{base model} for a new task obtains an interesting new twist. Training data for FT the model may arrive at the device sequentially, namely in successive data batches. Each batch of data may be exploited for FT the FM, or it may be discarded. This problem setting is particularly challenging for two main reasons. First of all, FT is a computationally resource-consuming process, and the device may not have enough resources to execute it at every time step. Second, the current data batch may be of limited added value to the existing model in terms of improving its performance, or the model may already be well-performing. In this setting, identifying \emph{when it is best to perform FT} of the model so as to improve its performance is an important problem. What makes the problem more interesting is that the performance of the model is revealed only after FT is performed, e.g., by measuring its performance on a held-out validation dataset.

The decision of when to update a large, deployed FM must be taken online, with very little labeled data from the new data distribution, which makes performance estimates noisy and thus unreliable. Also, the type and severity of distributional shift for the data, whether the feature distribution, the labels, or both, are usually unknown, so it is hard to predict how model performance will degrade. 
Finally, deriving an optimal fine-tuning policy requires trading off the explicit cost of updating (e.g., compute budget) against the expected performance improvement (e.g., accuracy), which is application- and task-specific and difficult to quantify precisely. 

We study the problem of optimal continual FT a FM on a single node wherein data from the same task (e.g., text classification) arrive in batches. At each time step, the learner can either fine-tune the model using the current batch, incurring a compute cost and a reward (e.g., performance improvement), or discard the data. In our setting, the state of the learning environment is characterized by the data distribution shift, the model’s performance, and the compute budget. 
\par An optimal fine-tuning strategy would adapt its update decisions to these evolving conditions. For example, allocating more training resources when a significant distribution shift is detected or when model performance deteriorates, while conserving resources otherwise. This adaptive behavior enables efficient use of the available budget by performing costly FT steps only when they are likely to yield meaningful performance improvements. Since the underlying data dynamics and the effect of FT on the model over time are not known in advance, the learner does not have access to the transition probabilities that govern how the state evolves. Consequently, the expected long-term rewards and costs cannot be computed analytically. To this end, we formulate this problem as a constrained Markov Decision Process (MDP).

\par We derive a policy that selects between fine-tuning and skipping, to optimize long-term reward, i.e., the model's performance, while respecting the compute budget constraint. It is based on the \emph{actor-critic} method on a Lagrangian version of the constrained MDP. We also study the special case where the performance improvement of FT a model can be predicted or estimated prior to decision; in this case the problem becomes a Dynamic Programming one.

The contributions of this paper are summarized as follows:
\begin{itemize}
\item We formalize the problem of deciding when to fine-tune as a budget-constrained MDP, and provide a principled framework for online FM adaptation under continuously arriving data batches.
\item We propose an \textit{actor–critic} reinforcement learning (RL) policy that jointly monitors performance degradation and data drift to trigger fine-tuning actions.
\item We study the special case where the expected performance gain from fine-tuning a FM can be estimated a priori, resulting in a Dynamic Programming formulation.

\end{itemize}


\section{Related work} \label{section:related}

\textbf{Parameter-Efficient Fine-Tuning (PEFT)}: PEFT adapts large pre-trained models by updating only a small subset of their parameters. $\mathrm{LoRA}$-based approaches~\cite{loraplus} reparameterize weight updates into low-rank subspaces, while freezing the original weights. Other PEFT strategies include adapter-based (e.g., inserting lightweight bottleneck modules into Transformer layers~\cite{houlsby}) and prompt-based (e.g.,  learning continuous prompts or prefix embeddings that steer model behavior without altering core weights~\cite{lester21prompt}). Despite rapid progress in PEFT, no single method consistently dominates across model sizes, tasks, or resource constraints~\cite{karimi21}. Our framework is method-agnostic, accommodating any fine-tuning approach.

\textbf{Continual PEFT}: Continual learning (CL) for large FMs has recently gained attention due to the challenge of mitigating catastrophic forgetting during sequential model updates. $\mathrm{Online-LoRA}$~\cite{Wei25} uses $\mathrm{LoRA}$ adapters and monitors loss plateaus, i.e., when the loss stops decreasing, to decide when to adapt. However, it does not formulate a cost-aware control problem on \emph{when} to adapt, nor does it optimize an explicit compute budget over time. 

Budget-adaptive PEFT methods~\cite{Wan25, ZhangQ23, Bhat25} address resource allocation along different axes. $\mathrm{OA-Adapter}$~\cite{Wan25} is a parameter-efficient CL method for LLMs that dynamically allocates adapter capacity across tasks and layers under a global parameter budget. In contrast, we assume a fixed PEFT method and optimize \emph{when} to invoke it over an incoming batch of data. $\mathrm{AdaLoRA}$~\cite{ZhangQ23} adaptively distributes a fixed low-rank parameter budget across weight matrices within a single downstream task. $\mathrm{PEARL}$ extends dynamic low-rank adaptation to a CL setting, where the rank of task-specific $\mathrm{LoRA}$ components is chosen based on the proximity of the current task to reference task weights in parameter space~\cite{Bhat25}. Both methods focus on \emph{how much} capacity to assign to each layer or task (rank selection / parameter count), assuming that FT is performed whenever a new task is encountered. In contrast, we keep the FT mechanism fixed and treat the \emph{timing} of updates as the primary decision variable.

Zliobaite et al.~\cite{Zliobait15} and Mahadevan et al.~\cite{Mahadevan24} explicitly study the cost of retraining. The former evaluates empirical adaptation methods, while the latter triggers retraining when the expected degradation cost of the model exceeds the compute cost of retraining. However, their approach relies on an \emph{offline-optimized} decision function trained on limited past data, which may not generalize to unseen data shifts. In contrast, we maintain a state-based online control process that continuously observes domain shift, performance, and resource usage and learns to make FT actions under a cumulative budget constraint. Additionally, we assume a pre-trained model which might perform well on unseen data and therefore not require retraining; a case not explicitly covered by \textit{CARA}~\cite{Mahadevan24} since not retraining the model incurs a staleness cost in their setting.

\textbf{Differentiation from existing works}: We address the underexplored question of \emph{when to fine-tune a large FM} in an online, streaming-data scenario. We do not rely on offline-optimized thresholds or staleness proxies; instead, we learn an online policy via a \emph{actor–critic} approach that optimizes long-term performance under budget constraints. Moreover, our formulation naturally addresses continual PEFT of large FMs, enabling adaptation in settings where retraining may provide diminishing returns, or may not be necessary at all.

\section{System Model and Problem Statement} \label{section:model}

\subsection{System model}

We consider an online setting in which a large pre-trained model parameterized by $\mathbf{\Theta_t}$ at time step $t$ is deployed at a resource-limited device and evolves over a time horizon $T$ in order to improve its performance on a downstream task, such as text classification. At each time step $t \in \{1, \dots, T\}$, a labeled dataset $\mathcal{D}_t = \{(x_i^t, y_i^t)\}_{i=1}^{n_t}$ arrives in a streaming manner. The learner must decide whether to update the model's parameters to improve its performance on this specific task,  under the resource limitations of the device.

Specifically, after the arrival of $\mathcal{D}_t$, and given the current model $\mathbf{\Theta}_{t-1}$, the learner has two options. If the learner decides to use $\mathcal{D}_t$ to fine-tune the model, then it uses $\mathcal{D}_t$ as input of the model and applies a certain fine-tuning method, yielding an updated model $
\mathbf{\Theta}_{t} = \mbox{FineTune}(\mathbf{\Theta}_{t-1}, \mathcal{D}_t)$,
where function $\mbox{FineTune}(\cdot,\cdot)$ denotes the effect of applying the selected FT method to the current model, $\mathbf{\Theta}_{t-1}$ with the current dataset $\mathcal{D}_t$. Our problem definition, to be detailed below, is agnostic to the specific FT method.
\begin{figure}[t!]
    \centering
    \includegraphics[width=\columnwidth]{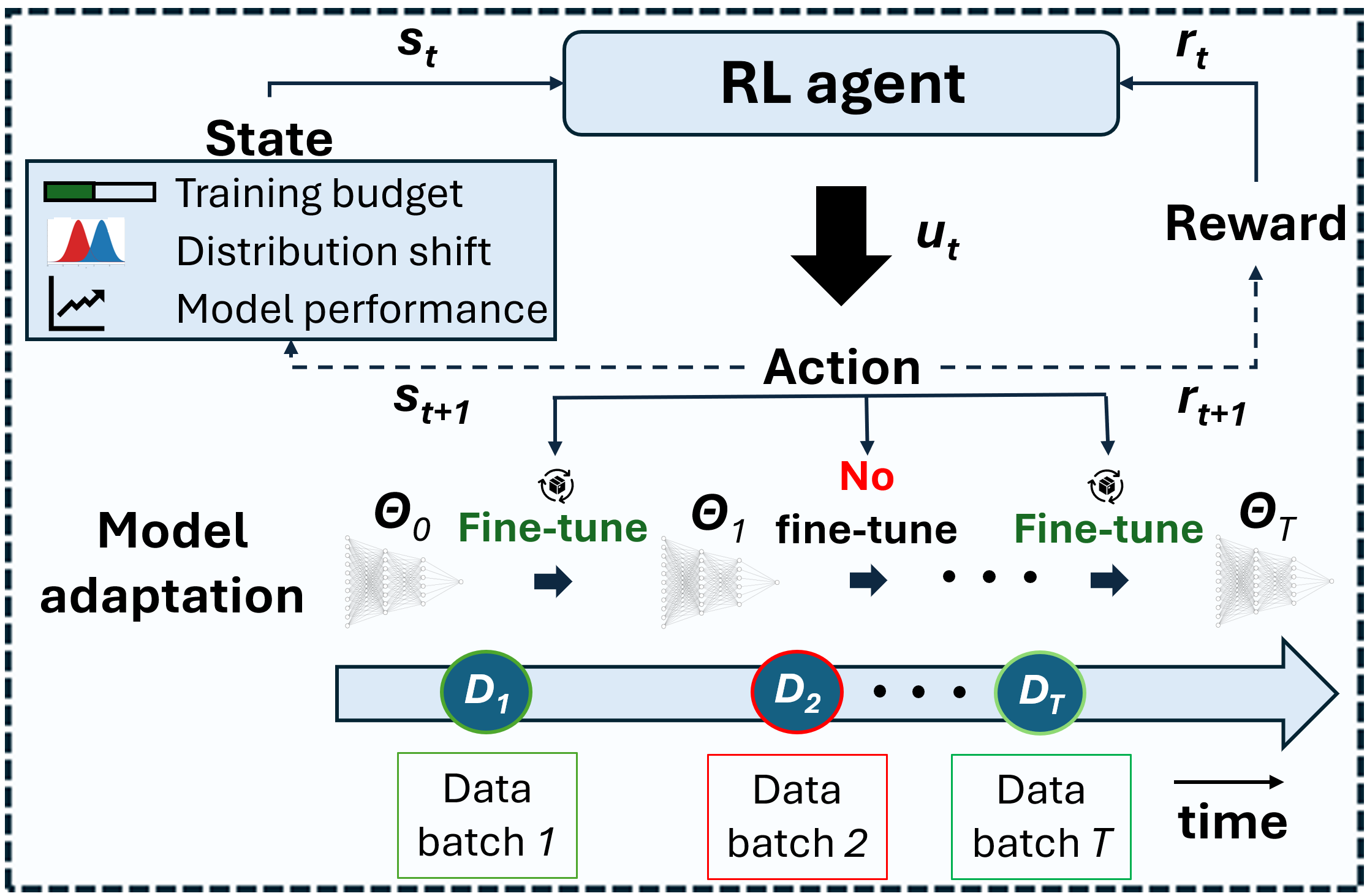}
    \caption{Our system model. At each step $t$, the RL agent observes the state $s_t$ (budget, data shift, and performance) and chooses whether to fine-tune the model on batch $\mathcal{D}_t$. The decision updates $\mathbf{\Theta}_t$ and yields reward $r_t$.}
    \label{fig:MAIN}
\end{figure}
\par Our system model (depicted in Fig.~\ref{fig:MAIN}) focuses on adapting the FM for a single machine learning task. As a concrete example, consider the task of text classification, where the goal is, for example, to determine whether a user review is positive or negative (sentiment analysis), or identify the purpose of a message (intent prediction). Our framework is general and can be applied to any single-task FT scenario.

We assume that there is a known cost $C(n_t)$ associated with the chosen fine-tuning method, which depends on the quantity of the data ($n_t = |\mathcal{D}_t|$), i.e., larger batches generally lead to higher cost, and may denote compute, energy or monetary costs. We denote by $B_{\max}$ the adaptation budget, which constrains the number of times when FT is performed over a finite time horizon $T$. After FT, the updated model is evaluated with respect to a performance metric, denoted as $a_t = a(\mathbf{\Theta}_t; \mathcal{D}_{\mathrm{ev}})$ (e.g., model accuracy for a classification task) on a fixed held-out dataset $\mathcal{D}_{\mathrm{ev}}$. If the learner decides not to perform FT, then it does not incur any cost, the model remains the same, i.e., $\mathbf{\Theta}_t = \mathbf{\Theta}_{t-1}$, and therefore the performance on $\mathcal{D}_{\mathrm{ev}}$ is the same as in the previous slot, i.e., $a(\mathbf{\Theta}_t; \mathcal{D}_{\mathrm{ev}}) = a(\mathbf{\Theta}_{t-1}; \mathcal{D}_{\mathrm{ev}})$.

\subsection{Problem statement}

\textbf{Action space}: Let $u_t \in \mathcal{A} = \{0,1\}$ denote the action of the learner, where $u_t = 1$ if the learner decides to fine-tune the FM, and $u_t=0$ if it decides not to fine-tune it.  

\textbf{State}: 
Intuitively, the learner’s decision process is stateful, as it must depend on key evolving quantities that reflect model performance, computational budget, and data distribution. Therefore, we formalize the system state at time $t$ as $s_t = (d_t, p_t, b_t)$, where 
\begin{equation}
p_t \triangleq
\begin{cases}
\displaystyle
\frac{1}{t}\sum_{i=1}^{t} a_i, & \text{if } t < w, \\[10pt]
\displaystyle
\frac{1}{w}\sum_{i=t-w+1}^{t} a_i, & \text{if } t \ge w,
\end{cases}
\end{equation}
denotes the moving average of the last $w$ application-specific performance metrics $a_i$, and it is updated online after each decision. The remaining budget is defined as 
\begin{equation}
b_t \triangleq B_{\max} - \sum_{i=1}^{t-1}  \mathbf{1}_{\{u_i = 1\}}\, C(n_i), 
\end{equation}
where $\mathbf{1}_{\{e\}}$ is the indicator function for the event $e$.
\par Finally, to quantify changes in the data distribution over time, we define a \emph{data shift score} based on the $\mathrm{KL}$ divergence:
\begin{equation}
d_t \triangleq D_{\mathrm{KL}}(P_t \| P^{\mathrm{hist}} ) = \sum_{c} P_t(c) \log \frac{ P_t(c)}{P^{\mathrm{hist}}(c)},
\end{equation}
where $P^{\mathrm{hist}}(\cdot)$ denotes the empirical historical distribution estimated over all past batches and $P_t(\cdot)$ the distribution of the current batch. The historical distribution can be updated online either as a cumulative average, $P^{\mathrm{hist}}_{t+1} = (t P^{\mathrm{hist}}_t + P_t)/(t + 1)$, or as an exponential moving average, $P^{\mathrm{hist}}_{t+1} = (1-\alpha) P^{\mathrm{hist}}_t + \alpha P_t$, which allows recent data to be weighted more heavily. By \emph{data distribution}, we refer to any relevant summary statistics that capture the dataset’s structure. A simple and common example is the \emph{class distribution}, where $P_t(c)$ is the fraction of samples in class $c$ within the current batch. Tracking the distribution shift is then straightforward: we maintain a histogram of class frequencies and compute the $\mathrm{KL}$ divergence between the historical and current distributions.

For non-classification tasks, such as translation, 
$P_t(\cdot)$ can represent distributions over token frequencies, n-grams, or embedding clusters. While these are high-dimensional, we can still maintain summary statistics (e.g., low-dimensional embeddings or histograms) to estimate $P_t(\cdot)$ and $P^{\mathrm{hist}}(\cdot)$
in a practical and computationally feasible manner.

\textbf{Cost}: The cost at each time $t$ is expressed as $c_t \;\triangleq\; \mathbf{1}_{\{u_t = 1\}}\, C(n_t)$.

\textbf{Transition:}  
The transition to the next state, $s_{t+1} = T(s_t, u_t)$, is stochastic and cannot be predicted or sufficiently estimated a priori, due to both the randomness of incoming data and the effect of fine-tuning. We characterize it as follows. \emph{If no fine-tuning is performed} ($u_t = 0$), the model parameters remain unchanged and no cost is incurred. A new batch $\mathcal{D}_{t+1}$ arrives, the data shift score $d_{t+1}$ is updated based on the new batch and the moving-average performance $p_{t+1}$ is computed based on the model of the previous slot, i.e., $\mathbf{\Theta}_{t-1}$,  resulting in the next state $s_{t+1} = (d_{t+1}, p_{t+1}, b_t)$. 
\par On the other hand, \emph{if fine-tuning is performed} ($u_t = 1$), the model is updated using the current batch $\mathcal{D}_t$, incurring a cost $C(n_t)$. After fine-tuning, the next batch $\mathcal{D}_{t+1}$ arrives, and the state variables are updated accordingly: the data shift score $d_{t+1}$ and moving-average performance $p_{t+1}$ are recomputed based on the updated model, yielding $s_{t+1} = (d_{t+1}, p_{t+1}, b_t - C(n_t))$. The next state depends only on the current state and action, making $(d_t, p_t, b_t)$ a sufficient state representation for our decision process under the modeling assumption that future data and performance depend on the past through these summary statistics.

\textbf{Reward}:  
The reward at each time step reflects the model's performance on a fixed held-out evaluation dataset $\mathcal{D}_{\mathrm{ev}}$, and is observed after the action is taken. If no fine-tuning is performed ($u_t = 0$), the reward is simply the current performance of the existing model, $r_t = a(\mathbf{\Theta_{t-1}}; \mathcal{D}_{\mathrm{ev}})$, while if fine-tuning is performed ($u_t = 1$), the reward is $r_t = a(\mathbf{\Theta_{t}}; \mathcal{D}_{\mathrm{ev}})$. In this way, the reward consistently reflects the impact of the chosen action on the model's performance.

Here, $a(\cdot \, ; \cdot)$ represents the metric of accuracy on a held-out validation dataset. Depending on the task, it can correspond to other standard metrics such as the $\mathrm{GLUE \,\, score}$ for natural language understanding tasks, $\mathrm{BLEU \,\, score}$ for language translation, or in general any other application-specific performance metric. This abstraction allows the framework to be applied flexibly across different tasks and evaluation criteria.

\textbf{Problem formulation}:  
We seek a stationary policy $\pi = \pi(u \,|\, s)$, such that the expected cumulative reward is maximized, subject to a given compute budget:
\begin{equation}
\begin{aligned}
&\max_{\pi} \quad && \mathbb{E}_\pi\Big[\sum_{t=1}^T r_t \Big] \\
&\text{subject to} \quad && \mathbb{E}_\pi\Big[\sum_{t=1}^T c_t\Big] \le B_{\max}, \\
& && s_{t+1}\sim T(\cdot\mid s_t,u_t), \;\; u_t\sim\pi(\cdot\mid s_t), \;\; s_1\ \text{given.}
\label{eq:formulation}
\end{aligned}
\end{equation}

The optimization problem (\ref{eq:formulation}) is a \textit{constrained Markov Decision Process} (MDP). Using this formulation allows the agent to optimally balance reward and budget under uncertainty, in contrast to naive heuristics such as always fine-tuning or fine-tuning based solely on a threshold on $d_t$ or $p_t$, which cannot adapt to the stochasticity in both data arrival and FT outcomes.  
\par Equivalently, introducing a dual multiplier $\lambda\ge0$, we form the Lagrangian objective:
\begin{equation}
\mathcal{L}(\pi, \lambda)\,\,\, = \,\,\,\mathbb{E}_\pi\Big[\sum_{t=1}^T \big(r_t - \lambda c_t\big)\Big] + \lambda B_{\max}.
\label{eq:lagrangian_objective}
\end{equation}

For any fixed $\lambda$, maximizing $\mathcal{L}(\pi,\lambda)$ with respect to $\pi$ is equivalent to solving an unconstrained MDP with a modified \emph{per-step reward}: $r'_t = r_t - \lambda c_t.$
We denote by $g(\lambda)$ the optimal value of this inner maximization:
\begin{equation}
g(\lambda)
=
\max_{\pi} 
\mathbb{E}_\pi\!\left[\sum_{t=1}^T (r_t - \lambda c_t)\right]
+
\lambda B_{\max}.
\label{eq:dual_function}
\end{equation}
The \emph{dual problem} is then $\min_{\lambda \ge 0} \, g(\lambda)$, and yields a policy that satisfies the budget constraint in expectation, or otherwise provides the best achievable trade-off between total reward and total cost. Intuitively, $\lambda$ acts as a penalty coefficient that balances model performance and resource expenditure: large values of $\lambda$ discourage fine-tuning actions by reducing their effective reward, whereas small values of $\lambda$ encourage them. 
The Lagrange multiplier $\lambda$ can be updated online using stochastic gradient ascent methods, which are known to converge asymptotically to policies achieving an optimal trade-off between reward and constraint satisfaction~\cite{Spoor25}.

\section{Reinforcement Learning solution} \label{section:solution}
\begin{algorithm}[t!]
\caption{Online budget-constrained fine-tuning}
\label{alg:online-budget-ft}
\begin{algorithmic}[1]
\REQUIRE actor \(\pi_\phi(u\!\mid\!s)=\sigma(f_\phi(s))\), critic \(V_\theta(s)\), \(\lambda\ge0\), per-step running cost \(\hat C\gets0\), horizon \(T\), total budget \(B_{\max}\), learning rates \(\eta_\phi,\eta_\theta,\eta_\lambda\), smoothing \(\beta_c\)
\FOR{$t=1,\dots,T$}
    \STATE Observe state $s_t=(d_t,p_t,b_t)$
    \STATE Compute action probability $\rho_t\triangleq\pi_\phi(1\mid s_t)=\sigma(f_\phi(s_t))$; sample $u_t\sim\mathrm{Bernoulli}(\rho_t)$
    \STATE Execute $u_t$: set $c_t \leftarrow \mathbf{1}_{\{u_t=1\}}\,C(n_t)$
    \STATE Receive reward $r_t$, next state $s_{t+1}$; update budget $b_{t+1}\leftarrow b_t-c_t$
    \STATE \textbf{Critic update (TD):} \[
        y_t \leftarrow r_t - \lambda c_t + V_\theta(s_{t+1}),\quad
        \delta_t \leftarrow y_t - V_\theta(s_t)
    \]
    \[
        \theta \leftarrow \theta + \eta_\theta \,\delta_t \,\nabla_\theta V_\theta(s_t)
    \]
    \STATE \textbf{Actor update (policy gradient):} \[
        \phi \leftarrow \phi + \eta_\phi\,\delta_t\,\nabla_\phi\log\pi_\phi(u_t\mid s_t)
    \]

    \STATE \textbf{Dual update:}  $\lambda \leftarrow \max\big(0,\ \lambda + \eta_\lambda(\hat C\cdot T - B_{\max})\big)$, where $\hat C \leftarrow (1-\beta_c)\hat C + \beta_c c_t$

\ENDFOR
\end{algorithmic}
\end{algorithm}

To derive an optimal policy $\pi = \pi(u \,|\, s)$, we employ a \emph{policy-gradient} method, specifically an \emph{actor-critic} algorithm. This choice is motivated by the continuous nature of the state space, which encodes model performance, data drift, and remaining budget, and by the stochasticity of state transitions. \emph{Actor--critic} methods are particularly effective in such settings, as they can jointly approximate the policy $\pi_\phi(u\mid s)$ (\emph{actor}) and the value function $V_\theta(s)$ (\emph{critic}) using differentiable function approximators such as neural networks. In our implementation, the actor is a logistic regression over the state features, while the critic is a linear regression.

The critic provides temporal-difference (TD) estimates of the Lagrangian-augmented return (with discount factor $\gamma=1$ due to the finite horizon), which guide the actor’s policy updates. The dual update uses an exponentially smoothed estimate $\hat C$ of the per-step cost with factor $\beta_c$ (e.g., $0.01$), which approximates the expected cost $\mathbb{E}_\pi[c_t]$ under the current policy. This approach reduces variance and stabilizes learning.  
The complete update procedure is summarized in Algorithm~\ref{alg:online-budget-ft}.

\section{An interesting special case} \label{section:specialcase}

We now study the following special case of the CMDP (\ref{eq:formulation}). Let $\mathbf{\Theta}_0$ be a pre-trained model  and $K$ the maximum times we are allowed to fine-tune the model. We assume that we know a priori the datasets $\mathcal{D}_t$, $t=1,\ldots,T$, and that the model's accuracy is obtained on a fixed dataset $\mathcal{D}_{\mathrm{ev}}$. This means that the accuracy can be obtained or estimated with sufficient reliability a priori. This scenario may arise either in an offline oracle setting, where all future datasets are known for planning purposes, or in cases where post-FT performance can be reliably predicted using for example large validation sets or surrogate models. Under this assumption, the FT decision problem becomes a Dynamic Programming one. We wish to find a FT policy, namely a sequence of times $J_m \triangleq \{j_1 < \ldots < j_m\} \subset  \{1, \ldots , T\}$, $m \leq K$ at which we fine-tune the model so as to maximize the cumulative accuracy over horizon $T$. 


\textbf{Accuracy depends on fine-tuning history.} In the general case, the evaluation accuracy at time $t$ depends on the model
parameters $\mathbf{\Theta}_t$, which themselves depend on the entire FT
history up to time $t$. Specifically, given a FT schedule
$J_m$, let $\mathcal{H}_t(J_m) \triangleq \{ j_i \in J_m : j_i \le t \}$
denote the set of fine-tuning times that have occurred up to time $t$.
The resulting model parameters $\mathbf{\Theta}_t$ are obtained by sequentially
applying the fine-tuning operator at the times in $\mathcal{H}_t(J_m)$, and the
corresponding accuracy is given by $a_t \;=\; a(\mathbf{\Theta}_t;\mathcal{D}_{\mathrm{ev}})$. We also define 
\begin{equation}
    A^*_T(m) \triangleq \max_{\substack{J_m = \{j_1 < \cdots < j_m\} }}
\; \sum_{t=1}^T a_t\big(\mathbf{\Theta}_t; \mathcal{D}_{\mathrm{ev}}\big),
\end{equation} 
the maximum total accuracy when fine-tuning is performed exactly $m$ times over time horizon $T$.

If $K=1$, we can choose either (\textit{i}) not to fine-tune at all during $T$ steps ($m=0$), with total accuracy in this case $A^*_T(0) = T a_0$; or (\textit{ii}) fine-tune once ($m=1$). In the second case, before fine-tuning at time step $t$, model $\mathbf{\Theta}_0$ is used, while after that the fine-tuned model $\mathbf{\Theta}_t$ is used. Then, $A^*_T(1) = \max_{j = 1,\ldots,T} [(j-1)a_0 + (T-j+1)a_j]$. Note that $\mathcal{H}_j(J_1) = \{j\}$.  The optimal time to fine-tune is the value of $j$ that attains the maximum in $A^*_T(1)$, and the optimal accuracy is $\max\{A^*_T(0), A^*_T(1)\}$.

If $K=2$, we can choose not to fine-tune at all with accuracy $A^*_T(0) = T a_0$, or fine-tune once with accuracy $A^*_T(1)$, or fine-tune twice. In the latter case, $A_T^*(2) = max_{1 \leq j_1 < j_2 \leq T} [(j_1 - 1)a_0 + (j_2 - j_1)a_{j_1} + (T - j_2 + 1)a_{j_2}].$ Then the optimal accuracy is $\max\{A^*_T(0), A^*_T(1), A^*_T(2)\}$. Note that the accuracy after the second fine-tuning time point, $a_{j_2} = a_{j_2}(\mathbf{\Theta}_{j_2}, \mathcal{D}_{j_2})$ depends on $\mathcal{H}_{j_2} = \{j_1\}$ as well. For $K>2$ the optimal accuracy is $\max_{m \in \{0,1,\ldots,K\}} A^*_T(m)$ with a total of $\sum_{m=0}^K \binom{T}{m}$ possibilities for fine-tuning times.

\textbf{Accuracy does not depend on fine-tuning history.} In another version of the problem, the accuracy $a_t$ at time $t$ may not depend on the current model $\mathbf{\Theta}_t$ (and thus, on fine-tuning history $\mathcal{H}_t$), but only on the current dataset $\mathcal{D}_t$. This is the case if $\mathcal{D}_t$ is large enough or if (close to) full FT is performed so that the impact of the pre-FT model on post-FT accuracy is small. Then, for each $t=1,\ldots,T$, we know a priori the accuracy $a_t$ on the evaluation set $\mathcal{D}_{\mathrm{ev}}$ if we fine-tune at time $t$. The problem is formulated as
\begin{equation}
    \max_{\{j_1, \ldots, j_r\}, r \leq K} (j_1-1)a_0 + \sum_{m=1}^r (j_{m+1}-j_m)a_{j_m}\,.
\end{equation}

The problem can be cast as a Dynamic Programming one. Let $V(t,k)$ be the maximum remaining total accuracy for times $t+1,\ldots,T$ when the model is fine-tuned at time $t$ and there are at most $k$ FT times remaining, for $t=0,\ldots,T-1$ and $k=0,\ldots,K$. For each $t$ if $k=0$, it is $V(t,0) = (T-t)a_t$ and we have accuracy $a_t$ for all remaining times. When $k \geq 1$, we have two options: either never fine-tune again and have total remaining accuracy $(T-t)a_t$ or choose the next FT time $j$, $t < j \leq T$. In this case, we use the current model for times $t+1,\ldots, j-1$ and get accuracy $(j-t-1)a_t$, and at time $j$ we fine-tune and the best future total remaining accuracy is $V(j,k-1)$. The Bellman equation is:
\begin{equation}
\begin{aligned}
V(t,k) = \max \Big\{ (T-t)a_t , \\
& \!\!\!\!\!\!\!\!\!\!\!\!\!\!\!\!\!\!\max_{j \in \{t+1,\ldots,T\}} \big[ (j-t)a_t + V(j,k-1) \big] \; \Big\},
\end{aligned}
\end{equation}
and we can recursively compute $V(s,k)$ for $s=T,T-1,\ldots 0$ and $k=0,1,...K$ until we find $V(0,K)$. For each pair $(t,k)$, we keep an index $i(t,k)$ which is equal to the value of $j$ that achieves the maximum at that step, or equal to null if the maximum is achieved with no FT.

\section{Numerical Results} \label{section:results}

We evaluate our approach on the $\mathrm{AG\,News}$ corpus, a standard text classification benchmark with four classes ($|\mathcal{Y}|=4$). The original training set is randomly shuffled and split into $90\%$ for adaptation and $10\%$ for evaluation. The evaluation split is held fixed throughout the experiment and used to measure model performance after each fine-tuning decision. As a base pre-trained model, we employ $\mathrm{RoBERTa}$~\cite{liu19roberta}. All experiments are repeated three times with different random seeds, and we report average results.

Data arrive in batches of fixed size $B = 200$ samples over $T = 200$ steps. Each batch corresponds to a FT decision, and the model is updated for a \emph{single epoch per batch} due to limited resources. We consider two strategies to create the batches: (\textit{i}) \textbf{All-labels:} Batches contain samples from all classes, randomly partitioned from the shuffled training set; and (\textit{ii}) \textbf{By-label:} Data are first partitioned by class and batches contain samples from a single class, resulting in $T/|\mathcal{Y}|$ batches per class. The latter setting induces distribution shifts over time and poses a more challenging adaptation problem.
\par We compare our method with baselines that cover a representative set of fine-tuning strategies under different computational budgets. The \textit{Always FT} baseline continuously FT the model on every incoming batch, using either full-parameter updates or \(\mathrm{LoRA}\) adapters, and serves as an empirical upper bound on achievable performance. The \textit{Never FT} baseline  never FT, deploying the pre-trained model as-is to provide a lower bound. Finally, the \textit{Random budgeted FT} baseline performs FT  either with full parameters or with \(\mathrm{LoRA}\), only for a limited number of updates, i.e., $50$ out of $200$ in total, and the FT steps are selected uniformly at random. Our proposed method operates under the same budget constraint as the \textit{random budgeted FT} baseline but adaptively determines \emph{when} to FT based on observed model performance and data drift.

\par
\begin{figure}[t]
    \centering
    \includegraphics[width=\linewidth]{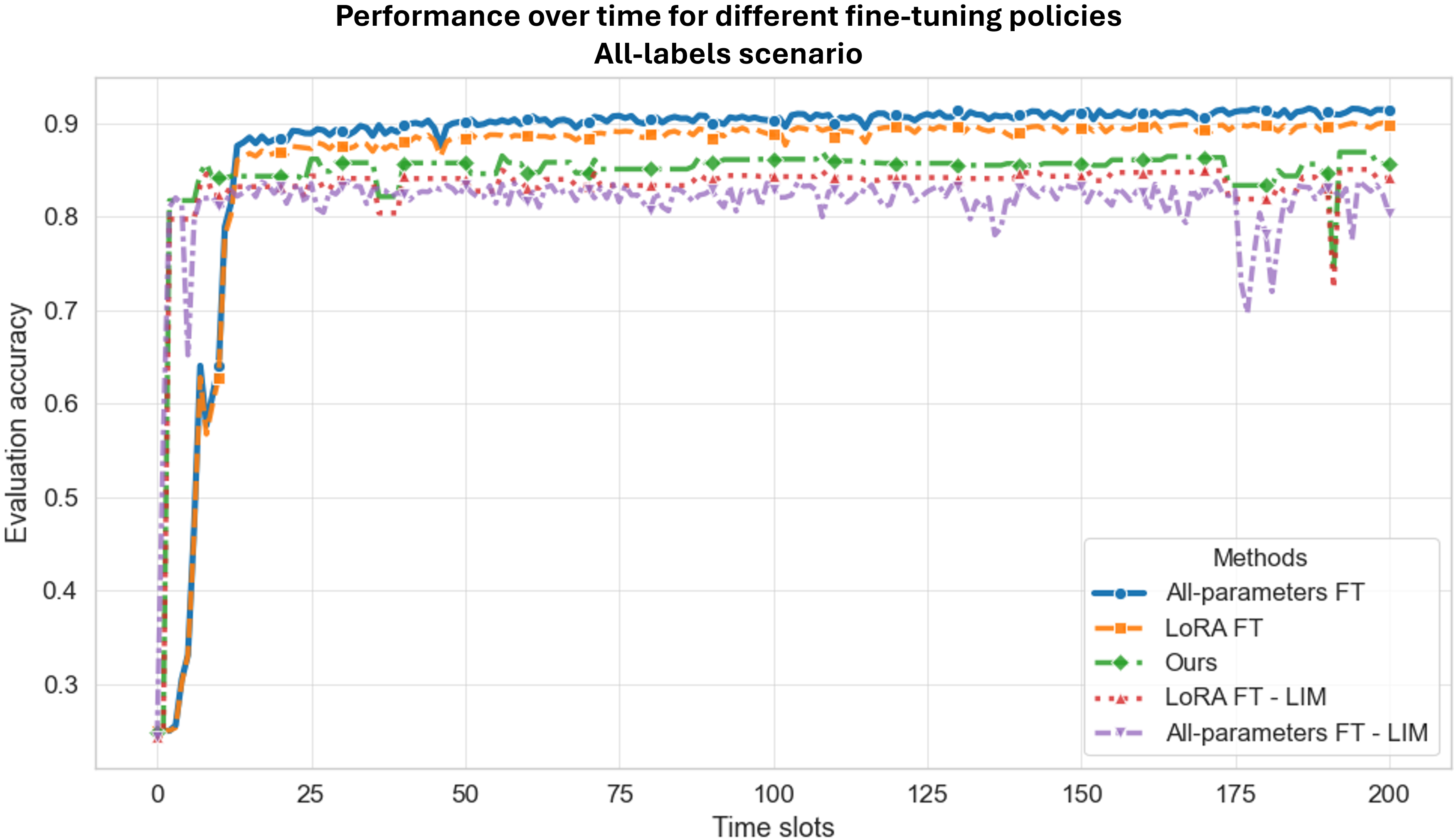}
    \caption{Accuracy over time in the \textit{all-labels} scenario. Our method achieves the best performance under the same update budget of 50 fine-tuning steps.}
    \label{fig:all_classes}
\end{figure}

 Figure~\ref{fig:all_classes} reports the evolution of accuracy over time on $\mathcal{D}_{\mathrm{ev}}$ for the \emph{all-labels} scenario. \emph{always FT} with full-parameter updates achieves the highest accuracy (peaking around $92\%$), serving as an \emph{empirical} upper bound. Under a limited (\emph{LIM}) budget of 50 updates, both random strategies (\emph{All-parameters FT–LIM} and \emph{LoRA FT–LIM}) exhibit noticeably lower and more unstable trajectories, converging to roughly $83\%$ accuracy. In contrast, our method maintains substantially higher accuracy under the same budget (around $89\%$), reducing the gap to the upper-bound \emph{full–FT} curves and clearly outperforming both random-budgeted baselines. Qualitatively similar trends are observed in the \emph{by-label} scenario; under a budget of 50 updates, our method achieves an accuracy of $69\%$, compared to $60\%$ for \emph{random budgeted FT} and $75\%$ for \emph{always FT}.

\par The advantage of our approach is also evident in Table~\ref{tab:Lora_table}, where $\mathrm{LoRA}$ is evaluated over different budget constraints. Our policy improves average reward (average evaluation accuracy over time) from $0.80$ to $0.85$ and best accuracy from $86.1\%$ to $89.2\%$ while using the exact same number of updates, and comes within $2.4\%$ of the always FT baseline while using only $25\%$ of the fine-tuning steps. These results highlight that the timing of updates is crucial for maximizing accuracy under strict computational budgets. The code to reproduce all experiments will be publicly released upon publication.

\begin{table}[t!]
\label{tab:Lora_table}
  \centering
  \caption{Comparison of fine-tuning policies with $\mathrm{LoRA}$ adapters.}
  \small 
  \newcolumntype{Y}{>{\centering\arraybackslash}X} 
  
  \begin{tabularx}{\columnwidth}{lYYYY}
    \toprule
    \textbf{Method} & \textbf{Avg. Reward} & \textbf{Best Eval. Acc. (\%)} & \textbf{\# FT steps}  \\
    \midrule
    Always FT             & 0.88 & 91.6 & 200  \\
    Never FT              & 0.61 & 64.1 & 0    \\
    Budgeted FT           & 0.80 & 86.1 & 50  \\
    \textbf{Ours}         & 0.85 & 89.2 & 50  \\
    \bottomrule
  \end{tabularx}
\end{table}

\section{Conclusion} \label{section:conclusion}

We studied the problem of deciding when to fine-tune a FM in the presence of limited compute budget. For the case when the outcome of FT is observed after the decision, we formulated the problem as a RL one, and we proposed a budget-constrained version of the \emph{actor-critic} algorithm to solve it. When the outcome of FT can be estimated before the decision, the problem admits a DP formulation.

A direct extension includes an enhanced action space, by incorporating several options for FT, each one with potentially different reward and cost. Then, the special case modeled by DP also warrants further investigation, with the identification of conditions under which the optimal policy admits simple intuitive policies. Finally, an interesting scenario would be if energy resources, and therefore the budget was renewed according to a certain process.

\section*{Acknowledgment}
The research project is implemented in the framework of H.F.R.I call “3rd Call for H.F.R.I.’s Research Projects to Support Faculty Members \& Researchers” (H.F.R.I. Project Number: 23767. Project Name: “Towards advancing the Mathematical and Computational Foundations for Digital Twins of Wireless Ad Hoc Networks”)

\end{document}